\documentclass[10pt,journal,compsoc]{IEEEtran}
\usepackage{amsmath,amsfonts}
\usepackage{amssymb}
\usepackage{algorithm}
\usepackage{array}
\usepackage[caption=false,font=normalsize,labelfont=sf,textfont=sf]{subfig}
\usepackage{textcomp}
\usepackage{stfloats}
\usepackage{url}
\usepackage{verbatim}
\usepackage{graphicx}
\usepackage{cite}
\usepackage[noend]{algpseudocode}
\usepackage{bm}
\usepackage{bbding}
\usepackage{colortbl}  
\usepackage{xcolor}
\usepackage{arydshln}
\usepackage[english]{babel}
\newcommand{\for}{\text{ }}

\begin{document}

\title{IMKGA-SM: Interpretable Multimodal Knowledge Graph Answer Prediction via Sequence Modeling}

\author{Yilin~Wen,~\IEEEmembership{}
        Biao~Luo,~\IEEEmembership{Senior Member,~IEEE,}
        ~and Yuqian Zhao % <-this % stops a space
%        ~and Yin Yang
\IEEEcompsocitemizethanks{\IEEEcompsocthanksitem Y.Wen, B.Luo and Y. Zhao are with the School of Automation, Central South University, Changsha, China.\protect\\
%Y. Yang is with the Hamad Bin Khalifa University, Doha, Qatar. \protect\\
E-mail: yilinwen510@gmail.com, biao.luo@hotmail.com, zyq@csu.edu.cn
%, yyang@qf.org.qa.
\IEEEcompsocthanksitem Corresponding Author: Biao Luo.}% <-this % stops an unwanted space
}

%\markboth{IEEE TRANSACTIONS ON KNOWLEDGE AND DATA ENGINEERING, VOL.~35, NO.~1, January~2023}%
\markboth{\tiny{This work has been submitted to the IEEE for possible publication. Copyright may be transferred without notice, after which this version may no longer be accessible}}%
{Shell \MakeLowercase{\textit{et al.}}: Bare Demo of IEEEtran.cls for Computer Society Journals}

\IEEEtitleabstractindextext{%
\begin{abstract}
Multimodal knowledge graph link prediction aims to improve the accuracy and efficiency of link prediction tasks for multimodal data. However, for complex multimodal information and sparse training data, it is usually difficult to achieve interpretability and high accuracy simultaneously for most methods. To address this difficulty, a new model is developed in this paper, namely Interpretable Multimodal Knowledge Graph Answer Prediction via Sequence Modeling (IMKGA-SM). First, a multi-modal fine-grained fusion method is proposed, and Vgg16 and Optical Character Recognition (OCR) techniques are adopted to effectively extract text information from images and images. Then, the knowledge graph link prediction task is modelled as an offline reinforcement learning Markov decision model, which is then abstracted into a unified sequence framework. An interactive perception-based reward expectation mechanism and a special causal masking mechanism are designed, which ``converts" the query into an inference path. Then, an autoregressive dynamic gradient adjustment mechanism is proposed to alleviate the insufficient problem of multimodal optimization. Finally, two datasets are adopted for experiments, and the popular SOTA baselines are used for comparison. The results show that the developed IMKGA-SM achieves much better performance than SOTA baselines on multimodal link prediction datasets of different sizes.\par
\end{abstract}

}

\maketitle

\IEEEdisplaynontitleabstractindextext

\IEEEpeerreviewmaketitle

\IEEEraisesectionheading{\section{Introduction}\label{sec:introduction}}

\IEEEPARstart{T}{he} knowledge graph is the technology and tool for carrying and representing background knowledge. It structures knowledge in the real world into entities and relations in the form of graphs and organizes them into networks. In a knowledge graph, knowledge data is represented in the form of triples $(h,r,t)$. Among them, $h$ is the head entity, $r$ is the relation connecting two entities, and $t$ is the tail entity. Knowledge graphs are used in various artificial intelligence tasks in different domains\cite{ref74}, such as named entity disambiguation \cite{ref59} in natural language processing\cite{ref75}, visual relation detection \cite{ref60} or collaborative filtering \cite{ref61}. However, it is well known that even state-of-the-art knowledge graphs are often incomplete (i.e., lack real facts or contain false facts). Therefore, machine learning algorithms aimed at addressing this problem attempt to infer missing triplets from observed connectivity patterns, a task is known as link prediction \cite{ref2}. For example, given a head entity and a relation $(h,r)$, predict a tail entity $t$. 

In order to solve the problem of link prediction, existing problems can be divided into four categories: deductive logic and rules, reasoning based on graph structure, knowledge graph embedded representation and deep neural network model. Rule-based reasoning methods, such as AMIE \cite{ref3}, AnyBURL \cite{ref4}, transform natural language queries into combinations of logical operators, express such queries through combinations, and then implement in a specific programming language to get the query. These methods are accurate and interpretable, but require experts to formulate a large number of inference rules, and have poor generalization ability for unknown rules. Reasoning based on graph structure has two features: one is the path feature, and the representative algorithm is PRA \cite{ref33}. The path features between nodes are extracted by graph traversal algorithm or random walk method, and the node connections are predicted by path features. Its characteristic is to provide path interpretability while reasoning, and the problem is that it is difficult to solve the problem because the reasoning nodes are not connected. The second is a graph-structure-based approach that utilizes a message-passing mechanism to extract the structural information of target entities and provide subgraph interpretability, and the representative algorithm is DeepPath \cite{ref8}. However, because the knowledge graph is usually very large, it is extremely complicated to traverse all the subgraph structures in the graph. The knowledge graph embedding representation method is to embed the high-dimensional and discrete data of the knowledge graph into a low-dimensional continuous vector space by designing a certain scoring function, and then representing the entities and relations as numerical vectors to calculate. Its representative model is the TransE type, for example, TransE \cite{ref1}, TransH \cite{ref5}, TransD \cite{ref66}, TransR \cite{ref6}. The recent research is bilinear models, e.g., RESCAL \cite{ref21}, DisMult \cite{ref22}, TuckER \cite{ref7}, and ComplEx \cite{ref16}. Its method is characterized by a shallow neural network, and the semantic representation of the knowledge graph is realized through a specific structure of the embedded space. The deep neural network model, e.g., CoKE \cite{ref63}, ConvE \cite{ref23}, is designed by designing entities and relations into query pairs, matching query pairs with entities relations, and obtaining inference similarity scores through deep neural networks to make inference judgments. Both the knowledge graph embedding model and the deep network model are regarded as neural network models, and the same point is that they both design a scoring function, and use the gradient backpropagation method for training in a data-driven manner. Its advantage is that its generalization performance is relatively better, and it effectively alleviates the problem of graph structure dimensionality disaster. Its disadvantage is that it only sees the similarity between input and output values, lacks interpretability, and performs single-step reasoning. In summary, as shown in Table~\ref{tab:s}, it is found that the methods based on logical deduction rules and graph structure are all symbol-based methods, which have better interpretability but poor generalization performance. Otherwise, based on knowledge graph embedding and deep neural network model, its generalization performance is better, but it lacks interpretability. Therefore, studying how to integrate symbolist and connectionist models is the key to obtaining an interpretable knowledge graph reasoning model.\par
\begin{table*}[t]
\centering
\caption{Summary of Existing Methods for Knowledge Graph Link Prediction}
\label{tab:s}
\begin{tabular}{cccccc}
\hline
reasoning algorithm         & logical rules                   & graph structure                    & knowledge graph embedding                     & deep neural network                    & reinforcement learning       \\ \hline
interpretability           & $\checkmark \checkmark \checkmark \checkmark$ & $\checkmark \checkmark$            & $\checkmark$                                  & -                  &    $\checkmark \checkmark \checkmark \checkmark$             \\
performance & $\checkmark$                                  & $\checkmark \checkmark$            & $\checkmark \checkmark \checkmark \checkmark$ & $\checkmark \checkmark \checkmark \checkmark$ & $\checkmark \checkmark$\\
robustness                 & $\checkmark \checkmark \checkmark \checkmark$ & $\checkmark \checkmark \checkmark$ & $\checkmark$                                  & $\checkmark$                                 &  $\checkmark\checkmark$ \\
scale                      & $\checkmark$                                  & $\checkmark$                       & $\checkmark \checkmark \checkmark$            & $\checkmark \checkmark$      &      $\checkmark \checkmark$            \\
expert experience          & dependence                                    & partial dependence                 & no dependence                                 & no dependence           & no dependence                      \\ \hline
\end{tabular}
\end{table*}
With the development of deep learning, the model structure of knowledge reasoning methods is becoming more and more complex. Because it is difficult for users to have an intuitive understanding of the parameters, structure and characteristics in such models, and they also have less understanding of the decision-making process and reasoning basis, it is difficult for users to trust the prediction results of the model. Therefore, in order to establish trust between users and reasoning models and balance the contradiction between model accuracy and interpretability, multi-hop reasoning methods are used to solve explainable knowledge reasoning\cite{ref67}. The rationale of the multi-hop reasoning method is to imitate the multi-hop thinking of human beings. A common approach is to apply reinforcement learning frameworks to multi-hop reasoning in knowledge graphs. Reinforcement learning is a model that has received a lot of attention in the past ten years and has been widely used in control \cite{ref71}, games \cite{ref72}, and robots \cite{ref73}. It models a learning process as a Markov process and trains the model by maximizing long-term cumulative rewards through the interaction between the agent and the environment. Modelling the knowledge map as a reinforcement learning process not only gets the result of reasoning, but also obtains the path of reasoning, and explains the reasoning of the knowledge graph through the reasoning path. The specific fusion method is to regard the knowledge graph as an environment, model the agent as a deep neural network, combine the advantages and disadvantages of symbolism and connectionism, and make the model have both the generalization performance and path interpretability of neural networks. Methods based on reinforcement learning such as DeepPath  \cite{ref8}, MINERVA \cite{ref9}, DIVINE \cite{ref10}, and AttnPath \cite{ref11}, however, generally have the shortcomings of slow convergence and low accuracy, and most of them are inferior to some traditional methods. The reason for this may due to the sparse rewards of reinforcement learning. Moreover, the sparse rewards, sparse data, and insufficient exploration of knowledge graphs make reinforcement learning more difficult and challenge in multimodal knowledge graph reasoning tasks  \cite{ref65}. Therefore, it is meaningful and promising to improve the accuracy of reinforcement learning in knowledge graph reasoning.

Recently, the cross-border application of Transformer  \cite{ref50} has attracted wide attention, and it has made breakthroughs in image classification  \cite{ref68}, semantic segmentation  \cite{ref69}, object detection  \cite{ref70} and other fields. Currently, Transformer has been employed as a pre-training model in offline reinforcement learning, e.g., Decision Transformer  \cite{ref17}, Trajectory Transformer \cite{ref46}, and Gato  \cite{ref18}, etc. These methods regard the data of reinforcement learning as a string of unstructured sequence data and train with supervised or self-supervised learning methods. It avoids the unstable gradient signal in traditional reinforcement learning and performs better in offline reinforcement learning. Deep reinforcement learning is a sequential process, therefore, the process of multi-hop reasoning is handled by state-of-the-art reinforcement learning sequence models, which may achieve better results than traditional reinforcement learning.\par

\begin{figure}[t]
\centering
\includegraphics[width=8cm]{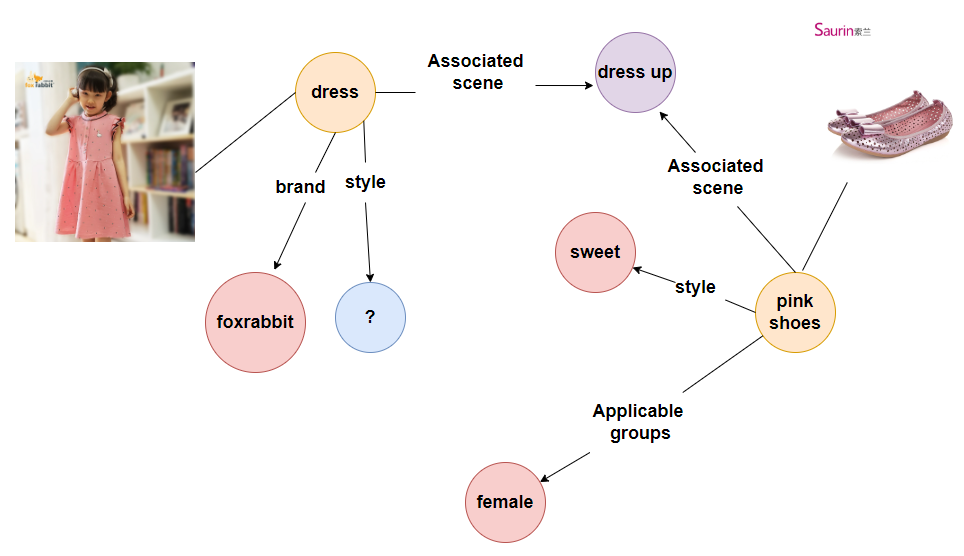}
\caption{When performing triple  $<shoes, style, ?>$ link prediction tasks, the answer is predicted based on triples $<dress, style, sweet>$ which is similar to the head entity image, and finally it is concluded that $?$ is $sweet$.}
\label{fig:first}
\end{figure}

For knowledge graph reasoning tasks that are complex and have the concept of multimodal data, the core idea of most existing knowledge graph reasoning algorithms is to reason by integrating existing triple structure knowledge, so knowledge of the entity is often ignored. However, information about entities themselves is usually beneficial for link prediction tasks, such as image and textual information. As shown in Fig.~\ref{fig:first}, for example, when performing the triple  $<shoes, style, ?>$ link prediction task, the answer is predicted based on the triple  $<dress, style, sweet>$ of the similar head entity image, and finally, answer $?$ is $sweet$. It is worth noting that the text information in the image also contains a lot of knowledge, especially when the knowledge graph is applied to the e-commerce field, the text in the product image is often the brand information of the product. Therefore,  to address multimodal explainable knowledge graph reasoning tasks with high efficiency and high performance, a new sequential model IMKGA-SM for reinforcement learning is developed, where a reward mechanism is designed based on perceptual interaction and fine-grained multimodal information extraction. \par
\par
\section{Related Works}
\subsection{Single-modal Knowledge Graph Reasoning}
Single-modal knowledge graph reasoning mainly revolves around relational reasoning. The AMIE \cite{ref3} and AMIE+ \cite{ref32} algorithms are derived from the early inductive logic programming system \cite{ref30}, emphasizing automatic rule learning methods. It has strong interpretability, however, all the above methods require expert design rules. Graph structure-based reasoning methods (e.g., path ranking algorithm \cite{ref33}) are also used to tackle such problems, which is interpretable but computationally intensive and time-consuming. The embedding-based methods include TransE \cite{ref1}, ConvE \cite{ref23}, RotatE \cite{ref27}, and TuckER \cite{ref7}. Each of these models is simple and the training speed is fast, but they are not interpretable. Reasoning methods based on neural networks include neural tensor networks \cite{ref34}, R-GCN \cite{ref35}, implicit ReasoNet \cite{ref36}, etc. They are able to learn to reason through implicit processing, which results in poor interpretability and unstable performance. In addition, there are typical reinforcement learning methods, e.g., DeepPath \cite{ref8}, MINERVA \cite{ref9}, RLH \cite{ref77}, GussuianPath \cite{ref78}, etc, which have better interpretability and inference performance than representation learning-based methods, but the disadvantage is that the effect is poor.\par

\subsection{Multimodal Knowledge Graph Link Prediction}
Compared with the single-modal knowledge graph link prediction task, the main contribution of the multi-modal knowledge graph link prediction task is to integrate multi-modal data knowledge into the plain text knowledge graph. In multimodal knowledge graph link prediction tasks, it is very necessary to combine the textual semantics of entities with multimodalities, such as semantics, vision, and hearing. IKRL \cite{ref37} is the first knowledge representation model that includes image information. For each entity, it learns two different representations based on triple structure information and image information, respectively. DKRL \cite{ref38} is a knowledge representation for fused descriptions. Similar to the IKRL model, the DKRL model also learns a representation based on structural information and a representation based on text descriptions for each entity. Based on the single-modal knowledge graph link prediction model TransE \cite{ref1}, an autoencoder is employed in TransAE \cite{ref14} to jointly encode visual information and text information to obtain the vector representation of entities. RSME \cite{ref15} is a multimodal knowledge graph reasoning model based on the traditional knowledge graph embedding model ComplEx \cite{ref16}. However, most of these multimodal approaches are uninterpretable and with low accuracy. \par

\subsection{Reinforcement Learning with Transformers}
In \cite{ref17}, Decision Transformer is proposed by modelling reinforcement learning tasks as a sequence framework transformer, based on which SQUIRE \cite{ref19} is employed to handle single-modal knowledge graph link prediction. However, these works are deficient in generalization and the reward information is underutilized. Based on Decision Transformer \cite{ref17}, Trajectory Transformer \cite{ref46} uses the beam search for model-based planning, while generating new trajectories is too complicated. Therefore, a simple random masking mechanism is proposed in this paper, which achieves the effect of data enhancement by randomly masking historical actions that have been generated in the past. Recently, Deepmind proposed a general agent, i.e., Gato \cite{ref18}, which made a further breakthrough in multimodal tasks. It is promising and potential to extend this model to multi-modal multi-hop reasoning.\par

\section{Methodology}
\begin{figure}[t]
\centering
\includegraphics[width=8cm]{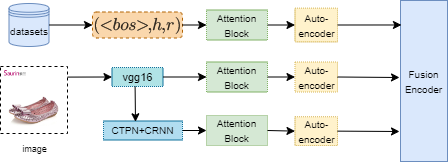}
\caption{The multimodal feature extraction module.}
\label{fig:2}
\end{figure}

\begin{figure}[t]
\center
\includegraphics[width=8cm]{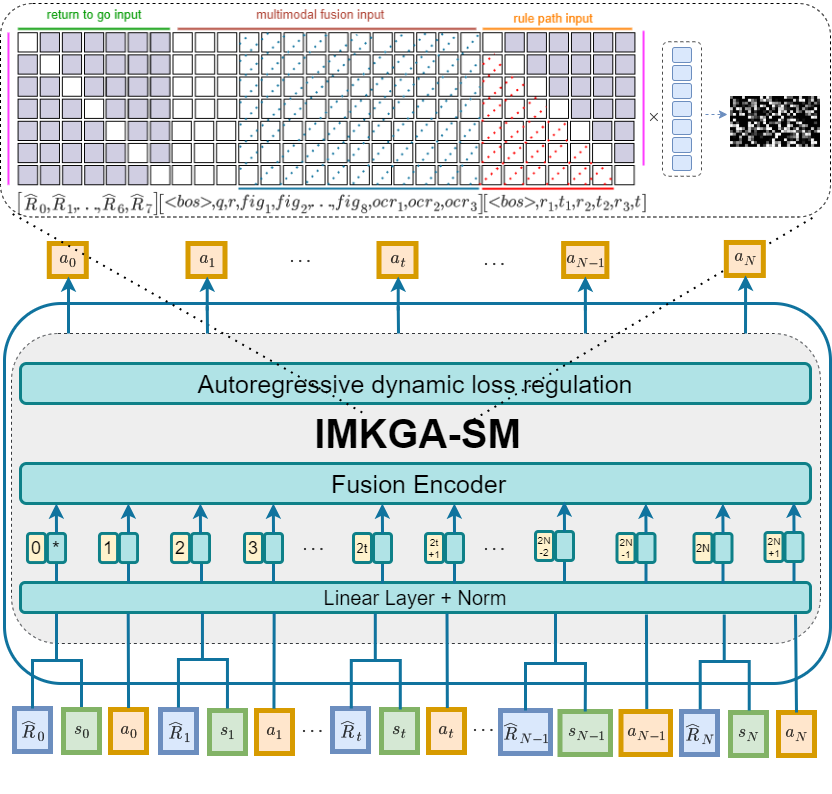}
\caption{Unified interpretable multimodal knowledge graph sequence framework.}
\label{fig:4}
\end{figure}

\begin{figure}[t]
\centering
\includegraphics[width=8cm]{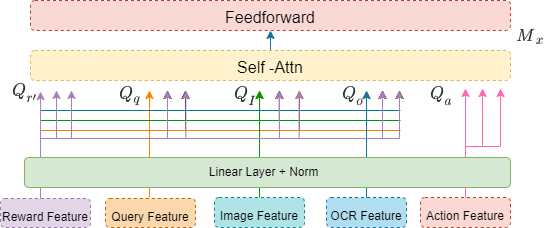}
\caption{The fusion encoder.}
\label{fig:3}
\end{figure}

\begin{figure}[t]
\centering
\includegraphics[width=8cm]{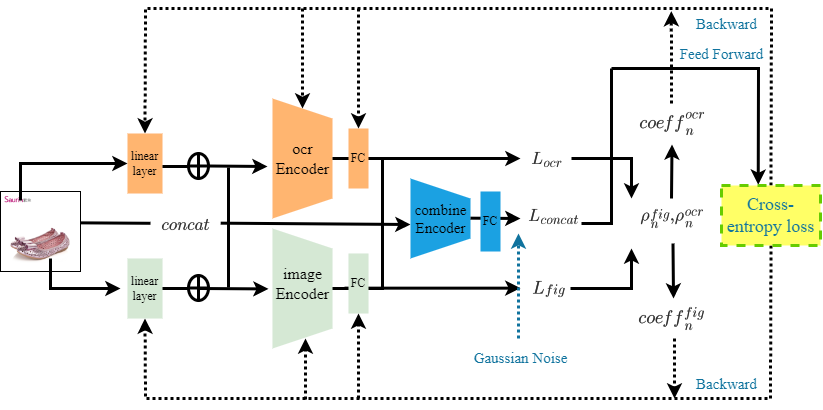}
\caption{Autoregressive dynamic loss regulation.}
\label{fig:5}
\end{figure}

In this section, the overall framework of IMKGA-SM is introduced, which treats the multi-hop reasoning problem as a sequence-to-sequence task derived from regression modelling trajectories and applies it to the task of multimodal link prediction. The hybrid transformer architecture of IMKGA-SM mainly includes five stacked modules. (1) The underlying multimodal feature extraction module, as shown in Fig.~\ref{fig:2}, aims to obtain basic structural information, image information, and text information in images from databases and images, and combine the three as a state feature. (2) The reinforcement learning sequence module, as shown in the bottom part of Fig.~\ref{fig:4}. The knowledge graph link prediction task is modelled as an offline reinforcement learning problem, which is then abstracted into a sequential framework. (3) The upper multimodal encoder (fusion encoder) module, as shown in Fig.~\ref{fig:3}, fuses the underlying features, reward features based on perceptual interaction, and action features through a self-attention mechanism. (4) The Mask mechanism module, as shown in the upper part of Fig.~\ref{fig:4}, includes three mechanisms to ensure the input and output of the encoder and prevent overfitting. (5) The loss function module, adopts an autoregressive self-adjusting mechanism to maximize the multi-modal performance, as shown in Fig.~\ref{fig:5}. \par

In the following subsections, each module of the IMKGA-SM will be analyzed and discussed in details. The multimodel feature extraction module and the reinforcement learning sequence architecture are developed in Subsections~\ref{p1} and \ref{p2}, respectively. The fusion encoder module is proposed in Subsection~\ref{p3}. The mask and loss function modules are designed in Subsection~\ref{p4} and \ref{p5}, respectively.\par
\par

\par

\par

\par
\subsection{Multimodal Feature Extraction Module}\label{p1}
In multimodal knowledge graph tasks with only single image data, most of the existing methods only learn simple image information. However, many visual scenes contain text with key information, so understanding text in images is crucial for downstream reasoning tasks, such as product brand, price, and consumer population. To jointly learn multimodal knowledge and inter-entity relations, knowledge in a single modality is extracted and combined into a multimodal transformer. In this paper, two modalities are considered: visual and textual, where text is extracted from image information. The multimodal part includes image input, text input in the image and query input (i.e. head entity, relation). Vgg16 pre-trained on ImageNet is used to process the head entity image information of the image input. Vgg16 consists of several vgg-blocks and three fully connected layers, and the vector output by the last fully connected layer is used as the image feature vector. \par

For image text input, OCR technology is used for image text extraction\cite{ref76}. Generally, OCR technology consists of two steps. (1) Text detection: locate the position of the text in the image. (2) Text recognition: identify the positioned text area, and convert the text area in the image into character information. In this paper, the CTPN method \cite{ref47} is adopted for text detection, and the CRNN method \cite{ref48} is adopted for text recognition. If the image information corresponding to the head entity is missing, $\emptyset$ is used instead. For the query input part, the knowledge graphs corresponding to head entities and relations are encoded to form a vector. Eq. \ref{eq.1} models an original multimodal feature $\phi$, which is specifically manifested as the fusion of structural information $(h,r)$, image information and text information.
\begin{equation}\label{eq.1}
\phi :G \times G \to G
\end{equation}
Here $h_{fig},h_{ocr},h, r \in G$. Then, let $*$ indicates a grouping operation, $h$ represents the structure embedding of the head entity, $h_{fig}$ represents the image embedding of the head entity, and $h_{ocr}$ represents the text embedding of the head entity after being extracted by the OCR method. Thus, as formalized in Eq. \ref{eq.2}, a characteristic entity $\tilde q$ of $\phi(h,h_{fig},h_{ocr})$ and $r$ is written as :
\begin{equation}\label{eq.2}
\tilde q = \phi \left( {h,{h_{fig}},{h_{ocr}}} \right) * r
\end{equation}
Multimodal fusion is widely used in the fields of computer vision\cite{ref79} and natural language processing\cite{ref80}. Since the currently most popular transformer framework is adopted as the core module of IMKGA-SM, according to the characteristics of the transformer, the number of parameters in the learning process largely determines the operation speed, so it is very necessary to process the input features of the transformer. Therefore, the module of the multimodal feature is a pre-train of the core transformer framework, which aims to filter out irrelevant or redundant features from the original data of features. Sepecifically, three self-attention blocks are used to receive the outputs of the original multimodal feature vector $\phi$, and three autoencoders are used to transfer them into a 14-dimensional vector in the end.\par

Specifically, first, the original multimodal feature $\phi_i$ passes through a fully connected feed-forward network to obtain different modal features ${\mu _i}$, which consists of two linear transformations and a ReLu activation function, via Eq. \ref{eq.3}.
\begin{equation}\label{eq.3}
{\mu _i} = conv\left\{ {ReLU\left[ {conv\left( {{\phi _i}} \right)} \right]} \right\},i= 1, \ldots ,L
\end{equation}
Then, as Eq. \ref{eq.4} shows, different modal features ${\mu _i}$ are passed through a Softmax layer in order to compute the attention of each modality $a_i$. 
\begin{equation}\label{eq.4}
{a_i} = Softmax \left( {{\mu _i}} \right),i= 1, \ldots ,L
\end{equation}
The sum of these attention weights $a_i$ multiplied by the original multimodal feature embedding $\mu_i$ is called self-attention $Q_\phi ^s$, formalized in Eq. \ref{eq.5}.
\begin{equation}\label{eq.5}
Q_\phi ^s = \sum\nolimits_{i = 1}^L {{a_i}{\mu _i}}  
\end{equation}
Therefore, $Q^s_{h}$, $Q^s_{h_{fig}}$, $Q^s_{h_{ocr}}$ of $h$, $h_{fig}$, $h_{ocr }$ are obtained respectively. $Q_\phi ^s$ is used as a query for the corresponding feature to calculate the attention weights guided by $Q_\phi ^s$ and put into the softmax layer. Finally, the weights are multiplied by the original modal features $\phi_k$ to get the filtered vector. Then, the output of the attention block is expressed as $g^\phi$ via Eq. \ref{eq.6}.
\begin{equation}\label{eq.6}
\begin{gathered}
  {p_k} = W\left[ {ReLu\left( {{W_s}Q_\phi^s} \right) \circ ReLu\left( {{W_x}\phi _k} \right)} \right], \hfill \\
  {s_k} = Softmax \left( {{p_k}} \right),k = 1, \ldots ,N, \hfill \\
  {g^\phi} = \sum\nolimits_{k = 1}^N {{s_k}\phi _k,\phi \in G}  \hfill \\ 
\end{gathered} 
\end{equation}
After feature $g^\phi$ is obtained, it is input into the autoencoder for dimensionality reduction. The final feature $h_\phi$ is shown in Eq. \ref{eq:7}, in which $h_{fig}$ is 8-dimensional, and $h_{ocr}$ is 3-dimensional.
\begin{equation}\label{eq:7}
{h_\phi } = \sigma \left( {W \cdot \phi  + b} \right)
\end{equation}
This part is used as the definition for the state $s$ of reinforcement learning, as shown in Fig.~\ref{fig:4}, which will be described in detail below.\par

\subsection{Reinforcement Learning Sequence Framework}\label{p2}
In this subsection, an offline reinforcement learning framework is developed for the knowledge graph link prediction task. Then, specific Markov triples are designed, and a reward expectation mechanism based on perceptual interaction is proposed. Finally, the whole reinforcement learning process is abstracted into a sequential framework, which is the core module of IMKGA-SM.

\subsubsection{Offline Reinforcement Learning Design}
The knowledge graph link prediction problem is modelled as a Markov decision process. Markov decision process tuple consists of a state $s \in S$, an action $a \in A$, a transition dynamic $P\left( {s'\left| {s, a} \right.} \right)$, and a reward function $r'$. $s_n$, $a_n$, and $r'_n$ are used to denote the state, action, and reward at time step $n$, respectively. A trajectory consists of a sequence of states, actions, and rewards: $\tau  = \left( {{s_0},{a_0},{r'_0},{s_1},{a_1},{r'_1}, \ldots,{s_N},{a_N},{r'_N}} \right)$. The purpose of reinforcement learning is to learn a policy that maximizes the expected return $E\left[ {\sum\nolimits_{n = 1}^N {{r'_n}} } \right]$ in the Markov decision process. In this paper, the process of generating a path for each triple link in the knowledge graph is regarded as a reinforcement learning segment. Since the data in this paper are fixed datasets, new data is difficult to be obtained through environmental interaction, so it is regarded as an offline reinforcement learning problem.\par

\subsubsection{Track Representation}
The inference accuracy of reinforcement learning-based inference methods is usually much lower than that of traditional TransE-based methods. This is because the amount of data for offline reinforcement learning is very limited, and the rewards for the knowledge graph link prediction problem are sparse, resulting in serious decision bias in reinforcement learning. Therefore, they are not suitable for direct transfer to the task of multimodal knowledge graph link prediction. To address this problem, a new reinforcement learning sequence framework with perceptual interaction expected reward mechanism is proposed in this subsection. Different from traditional reinforcement learning, the reward setting here is the expected reward in the future, that is, the maximum reward value expected to be obtained in the current state. There are two main differences: (1) An expected reward mechanism is proposed to eliminate the sparsity of rewards by incorporating the perceptual similarity of knowledge graph entities. (2) The multi-modal perception interface is introduced into the decision transformer framework for the first time, making full use of multi-modal features.

Through the previous pre-training process, the knowledge graph link prediction process is transformed into a Markov decision process. Its purpose is to find a path to the target entity, which means that the pathfinding process makes multi-hop reasoning interpretable. Therefore, the knowledge graph link prediction task is modelled as an offline reinforcement learning task, which is then transferred to a sequential framework for solving. Similar to offline reinforcement learning, the Markov triplet $<\hat{R},S,A,P_r>$ of IMKGA-SM is defined as follows.\par
State $s_k$: For the knowledge graph triple $<h,r,t>$ in the dataset, the state of IMKGA-SM is denoted as $s=(query, h_{fig},h_{ocr}) \in S$. Among them, $S$ is the state space, and $query=(<bos>, h, r)$ is the query of triples, representing the beginning, head entity, and relation, respectively. $h_{fig}$ and $h_{ocr}$  represent the image embedding of the head entity and the text embedding in the image, which are obtained by Eq. \ref{eq:7}.\par
Action $a_k$: The action space for a given $s_n$ is the set of all entities, relations and $<eos>$. The purpose is to infer the path from the head entity $h$, relation $r$ to the tail entity $t$ by generating the action output, and $a_k$ is the $k$-th action represented by the $k$-th token of the path generated by the rule. Here, AnyBURL \cite{ref4} is used as the rule-based method to find the path between $h$ and $t$, $r\left( {h,t} \right) \to {r_1}\left( {h,{t_1}} \right) \wedge {r_2}\left( {{t_1},{t_2}} \right) \wedge  \ldots  \wedge {r_n}\left( {{t_n},t} \right)$, decomposing a single relation into a combination of multiple relations and entities. If the model makes an error during the prediction process, the inferred entity or relationship does not conform to the corresponding attribute. Then the rule path is modified to remember only the last target entity. Taking three-hops as an example, $A = \left( {0,\emptyset,t,0,0,0,0} \right)$ is modified as a set of actions.\par
Transition $P_r$: The transition function $P_r$ is set to map the current state $s_n$ to the next state $s_{n+1}$. Formally, ${P_r}:S \times A \to S$ is defined as ${P_r}\left( {{s_n},{A_n}} \right) = {P_r}\left( {query,{h_{fig}},{h_{ocr}},{a_0}, \ldots ,{a_{n - 1}}} \right)$. When entering the next state, the actions of the previous step are added to the previous state as history to realize the state change. The Mask mechanism  \uppercase\expandafter{\romannumeral1} does the specific step. Therefore, unlike the traditional Decision Transformer \cite{ref17}, the state transition mechanism ${P_r}\left( {{s_n},{A_n}} \right)$ is designed to let the model focus on the state, actions and return-to-go of the previous steps, thereby improving the policy.\par
Return-to-go $\hat R$: Since the purpose of knowledge graph link prediction is to reason about the tail entity answer, it is impossible to know whether the reasoning is successful unless the last step of the reasoning path is reached. Therefore, there is a phenomenon of sparse rewards in the reinforcement learning sequence method when solving the knowledge graph link prediction task. In order to solve these problems, a reward expectation mechanism based on perceptual similarity is designed to learn interactively by taking the expected reward of the current state as input.\par
The definition of $\hat R$ is the maximum reward expected in the current state, so after making an action, the value of the next $\hat R$ will decrease (or increase) due to the reward from the previous action, ${\hat R_n} = \sum\nolimits_{n' = n}^ T {{r'_n} } $. So $\hat R$ changes with state and action. After obtaining the initial triple $query=(<bos>,h,r)$ and obtaining the corresponding reasoning path according to the rules, the return-to-go corresponding to each step $\hat R_n$ is obtained. $\tau(h,r,t)$ is defined as a set of triples in the dataset, and $\psi(<bos>,r_1,t_1,r_2,\ldots,r_n,t_n)$ is the path obtained by the rule that satisfies $\tau$. All dataset entities and relations are stored in collections $\mathbb{E}$ and $\mathbb{R}$. Specifically, the generation steps of $\hat R$ are as follows:\par

(1) At the initial value ($n=0)$, when no action is taken, the maximum expected reward of the task $\hat R_0$ is expected to be able to reach the target entity, which is a fixed value defined in Eq. \ref{eq.8}.

\begin{equation}\label{eq.8}
{\hat R_0} = {r'_{good}}
\end{equation}

(2) When $n=1$, the first action is $<bos>$, which means the beginning. As Eq. \ref{eq.9} shows, the first return-to-go $\hat R_1$ is defined according to whether the correct tail entity is finally successfully inferred. Here, $r'_{good}$ represents a positive constant and $r'_{bad}$ represents a negative constant.

\begin{equation}\label{eq.9}
{\hat R_1} =  \begin{gathered} \begin{aligned}
\begin{cases}
  {r'_{good}},&\for \psi \left( {{t_n}} \right) = t, \hfill \\ 
  {r'_{bad}},&\for \psi \left( {{t_n}} \right) \ne t \hfill \\ 
\end{cases}\end{aligned}
\end{gathered}  
\end{equation}

(3) For each action with $n>1$, return-to-go in $n$-th step is defined as Eq. \ref{eq.10}, in which a base penalty $r'_{step}$ (negative) is generated since as few hops as possible are desired to be used. When the current action $a_n$ is the entity $t_n$, an additional reward $r'_{{add}_n}$ (positive) will be generated.

\begin{equation} \label{eq.10}
{r'_n} = {r'_{step}} + {r'_{{add}_n}}, n > 1
\end{equation}

 (4) Define $r'_{{add}_n}$ as:
 \begin{equation}  \label{eq.11}
{r'_{{add}_n}} =  \begin{gathered}\begin{aligned}
\begin{cases}
{\hat R_1} \times sim\left( {{f_{{t_n}}},{f_t}} \right) & \for {a_n} \in \mathbb{E}, \hfill \\
  0,& \for {a_n} \in \mathbb{R} \hfill \\
  \frac{1}{2}{{\hat R}_1},& \for e{m_{{t_n}}} = \emptyset |e{m_t} = \emptyset  \hfill \\ 
 \end{cases}\end{aligned}
\end{gathered} 
\end{equation}
Here $sim(\cdot)$ denotes the cosine similarity computed with $sim\left( {u,v} \right) = \frac{{{u^T}v}}{{\left\| u \right\| \cdot \left\| v \right\|}}$, ${f}_{t_n}$ and ${f}_{t}$ are the image embedding of the current and target entities, respectively. From Eq. \ref{eq.11}, it is noted that the more similar the generated entity action is to the target entity, the greater the reward for that action should be. If the current action is the relation $r'_n$, or the target entity or the current recommendation entity has no image information, the additional reward $r'_{{add}_n}$ is 0. If the entity action $a_n$ has no image information or the target entity $t$ has no image information, the similarity value takes an intermediate value of 0.5.

(5) An additional penalty $r'_{bad}$ will be imposed if the recommended action does not conform to the attribute (entity or relation) that should be recommended. When performing action $a_n$ in the current state $s_n$, Eq. \ref{eq.12} defines the next Return-to-go input $\hat R_n$ as the previous step's Return-to-go $\hat R_{n- 1}$ minus the reward $r'_{n-1}$ caused by action $a_{n-1}$.
\begin{equation}\label{eq.12}
{{\hat R}_n} =\begin{gathered}\begin{aligned}
\begin{cases}
  {{\hat R}_{n - 1}} - {r'_{n - 1}} - {r'_{bad}},& \for {a_{n - 1}} \notin \mathbb{E},{a_{n - 1}} \notin \mathbb{R} \hfill \\
  {{\hat R}_{n - 1}} - {r'_{n - 1}},& \for \text{others} \hfill \\ 
  \end{cases}\end{aligned}
\end{gathered} 
\end{equation}

(6) Repeated iteration until the end of the round, if it is less than three hops, in order to ensure the same embedding length, the last $\hat{R}$ completion vector will be used to length 7. The trajectory $\tau$ is expressed in Eq. \ref{eq.13}.
\begin{equation}\label{eq.13}
\tau=\left( {{{\hat R}_1},{s_1},{a_1},{{\hat R}_2},{s_2},{a_2 }, \ldots ,{{\hat R }_{N,}}{s_N},{a_N}} \right)
\end{equation} 

\subsection{Fusion Encoder Architecture}\label{p3}
Sequences of token embeddings from the three modes, return-to-go, state, and action, are concatenated and fed to the transformer. Different from the positional embedding of the traditional transformer \cite{ref50}, a time step (return-to-go, state) shares the same positional embedding and the position of them are processed as a complete sequence. Therefore, the process of positional embedding is expressed as Eq. \ref{eq.14}. Here $X^U_{pc}$ is the projected embedding vector, $C$ is the concat operation, $U_{pos}$, represents the position embedding corresponding to the embedding layer, and $U = \left\{ {C\left( {\hat R,s} \right),a} \right\}$.
\begin{equation}\label{eq.14}
\begin{gathered}
  X_\tau^{C(\hat R,s)} = X_{pc}^{C(\hat R,s)} + C{\left( {\hat R,s} \right)_{pos}} \hfill \\
  X_\tau^a = {X^a_{pc}} + {a_{pos}} \hfill \\ 
\end{gathered} 
\end{equation}
To avoid increased computational complexity due to long concatenated sequences, Eq. \ref{eq.15} models $\bar X_\tau^{C(\hat R,s)}$ by adding an embedding linear layer $LN$ for each modality such that the original input is projected to the embedding dimension, followed by layer normalization $Sigmoid$.
\begin{equation}\label{eq.15}
\begin{gathered}
  \bar X_\tau^{C(\hat R,s)} = Sigmoid\left( {LN\left( {X_{pc}^{C(\hat R,s)}} \right)} \right) \hfill \\
  \bar X_\tau^a = Sigmoid\left( {LN\left( {X_{pc}^a} \right)} \right) \hfill \\ 
\end{gathered} 
\end{equation}
These tokens are processed by an encoder model that predicts future action tokens via autoregressive modelling. Since the multi-hop inference is a fixed-length sequence, the transformer's encoder structure is selected, which consists of $L$ stacked blocks. As shown in Fig.~\ref{fig:3}, each block mainly includes two types of sublayers: multi-head self-attention $MHA$ and fully-connected feed-forward network $FFN$. The transformer model contains many parameters including $W_V$, $W_Q$, $W_K$ matrices, and the values in each stack and head are designed. As formalized in Eq. \ref{eq.16}, multi-head attention mechanism $Attn$ is introduced to defined ${head}_i^{M_r'}$. Here, $r'$ and $a$ represent return-to-go and action features in reinforcement learning sequence framework, $q$ represents knowledge graph query $(<bos>,h,r>)$ feature, and $I$ and $O$ represent the image and OCR features in multimodal feature extraction module. 
\begin{equation}\label{eq.16}
\begin{aligned}
 head^{{M_r}} = & Attn\left( {{x^{r}}W_Q^{r}} \right.,\left[ {{x^{r}}W_K^{r},{x^{q}}W_K^{q},{x^{I}}W_K^{I},{x^{o}}W_K^{o}} \right], \hfill \\
  &\left. {\left[ {{x^{r}}W_V^{r},{x^{q}}W_V^{q},{x^{I}}W_V^{I},{x^{o}}W_V^{o}} \right]} \right) \hfill \\ 
\end{aligned}
\end{equation}
The calculation method of  ${head}_i^{M_q}$, ${head}_i^{M_I}$ and ${head}_i^{M_o}$ is similar to that of ${head}_i^{M_r}$, where ${head}_i^{M_a}$ is redefined via Eq. \ref{eq.17}.
\begin{equation}\label{eq.17}
head^{{M_a}} = Attn\left( {{x^{a}}W_Q^{a},{x^{a}}W_K^{a},{x^{a}}W_V^{a}} \right)
\end{equation}
Hence, Eqs. \ref{eq.18} and \ref{eq.19} model the hidden state of the encoder layer $l$.\par
\begin{equation}\label{eq.18}
\bar X_l^U = MHA\left( {LN\left( {\bar X_\tau ^U} \right)} \right) + X_{l - 1}^U
\end{equation}
\begin{equation}\label{eq.19}
X_l^U = FFN\left( {LN\left( {\bar X_l^U} \right)} \right) + \bar X_l^U
\end{equation}
Next, with the mask shown in Fig.~\ref{fig:4}, the encoder only focus on previous labels $a<k$ of the current return-to-go, multimodal fusion input and output paths. The specific details of the mask will be described in the next subsection. \par
\subsection{Mask Mechanism Design}\label{p4}
In the link prediction task of the recommendation system, due to feature redundancy, lack of sufficient training data and complex model design, the recommendation system is extremely prone to the one-epoch phenomenon, that is, the over-fitting phenomenon. Therefore, three mask mechanisms are designed to overcome the overfitting phenomenon, and they are also used to solve the input and output requirements of the reinforcement learning framework established in Subsection \ref{p2}.\par
Mask mechanism  \uppercase\expandafter{\romannumeral1}: As shown in the shaded area in Fig.~\ref{fig:4}, Mask mechanism  \uppercase\expandafter{\romannumeral1} is used to ensure the input and output of the transformer and realize the Markov decision process. Through step-by-step prediction, the path is predicted sequentially to obtain the final target entity. When predicting the next action, the previous action history will be added to the state to achieve state transition, that is, the real result of the previous step will be used as input to predict the output of the next path token. In this way, the context information is effectively used by the model, thereby ensuring the accuracy of the model, so that the final result will not have a large error due to a one-step error.\par
Mask mechanism  \uppercase\expandafter{\romannumeral2}: As shown in the blue dots in Fig.~\ref{fig:4}, Mask mechanism  \uppercase\expandafter{\romannumeral2} is used to solve the problem of model overfitting. After multimodal information dimensionality reduction, data of training features is  sparse, which easily leads to  fast immature convergence of the model, resulting in overfitting. Therefore, it is necessary to perform data dropout on this part of the embedded input. The double data dropout mechanism is introduced for data enhancement, which is conducive to retaining the original high-quality samples as much as possible. Specifically, for a given sequence, the data loss scheme is enabled with a certain probability $p_k$, and when applying the data loss scheme, tokens in the sequence are randomly masked with a certain probability $p_m$.\par
Mask mechanism \uppercase\expandafter{\romannumeral3}: As shown in the red dot in Fig.~\ref{fig:4}, the Mask mechanism \uppercase\expandafter{\romannumeral3} is used to make the model generate more new trajectories by itself, which makes the trajectories generated by the transformer in the past randomly masked out for the next action prediction task such that the model gradually learns from the self-generated trajectories. Mask mechanism \uppercase\expandafter{\romannumeral3} is simple and easy to implement, and it does not add any additional computational cost and parameters.\par
In the masking mechanism, as formalized in Eq. \ref{eq.20}, multiplying each token $x_i \in x$ by the mask gets its autoregressive log maximum likelihood. Here $\eta$ represents the mask ratio.
\begin{equation}\label{eq.20}
\log p\left( x \right) = \log \prod\limits_{i = 1}^n {p\left( {{x_i}\left| {\left[ {I\left[ {{m_j}  \leq \eta } \right] \cdot {x_j}} \right]} \right._{j = 0}^{i = 1}} \right)} ,\eta  \in \left[ {0,1} \right]
\end{equation}
Mask mechanism  \uppercase\expandafter{\romannumeral1} adopts a complete mask, so $\eta=1$. Mask mechanism  \uppercase\expandafter{\romannumeral2} and \uppercase\expandafter{\romannumeral3} are random masks, which are randomly masked according to a certain probability value $m_j$.\par
\subsection{Loss Function Design}\label{p5}
The multimodal information of entities (features of images and text features in images) is expected to enhance learning through fusion. However, experiments have found that after incorporating multiple modalities, the model will suffer from a lack of optimization, which is caused by the dominance of one mode in some scenarios. For example, image information dominates when inferring relations is relevant to tasks such as ``colour, item type". When reasoning about relations is a relevant task like ``brand", textual information in images dominates. Therefore, inspired by \cite{ref40}, a dynamic gradient adjustment mechanism is introduced to train three models separately, taking two modes and their concat as three inputs. By monitoring the contribution of each mode to the learning objective, each mode optimization is adaptively controlled, thereby alleviating the imbalance of mode optimization.

Three transformer's encoders,  represented by $Enc\left(\cdot\right)$, are accepted three modal features. When decoding ${\psi _k}: = \left( {{r_1},{t_1}, \ldots ,{r_n}t, < eos > } \right)$, $Softmax$ is used in Eq. \ref{eq.21} to calculate the distribution $p_i^\chi$, where $\chi  \in \left( {fig,ocr} \right)$, $b$ is the bias of the prediction model \cite{ref41}, and the addition of $b/2$ is used as a bias compensation of single-modal prediction.
\begin{equation}\label{eq.21}
\begin{aligned}
  p_i^\chi  = & \prod\limits_{k = 1}^{\left| \psi  \right|} {Soft\max \left( {MLP} \right.\left( {Enc\left( {\psi _n^\chi \left( {{\theta ^\chi },x_n^\chi  + \frac{b}{2}} \right)} \right.} \right.}  \hfill \\
 & {\left. {\left. { \cdot \left. {W_n^\chi } \right)} \right)} \right)_k} \hfill \\ 
\end{aligned} 
\end{equation}
In the same way, the distribution of concat feature $p_i^{concat}$ is shown in Eq. \ref{eq.22}. 
\begin{equation}\label{eq.22}
\begin{aligned}
  p_i^{concat} = &\prod\limits_{k = 1}^{\left| \psi  \right|} {Softmax\left( {MLP\left( {Enc\left( {\left( {\psi _n^{concat}\left( {{\theta ^{concat}}} \right.} \right.} \right.} \right.} \right.} , \hfill \\
  &\left. {x_{concat}} \right){\left. {\left. {\left. { \cdot W_n^{concat}} \right)} \right)} \right)_k} \hfill \\ 
\end{aligned} 
\end{equation}
As Eq. \ref{eq.23}-\ref{eq.25} shows, a cross-entropy loss is used, where $\varepsilon$ is a label smoothing hyperparameter ranging from 0 to 1 to avoid overfitting. A single-modal image feature, a single-modal OCR feature, and a multi-modal feature which is defined as the concat of both are fed respectively to compute three different losses. At the same time, since the previously designed mask mechanism shields some features, it is also necessary to exclude the loss caused by the mask token. Also, to prevent the model from giving higher scores to shorter paths, the sum of log-likelihoods divided by the length of the path is used.
\begin{flalign}\label{eq.23}
   {L_{fig}} =  - \frac{1}{{\left| {p_{mask}^{fig}} \right| \cdot N}}\sum\limits_{i = 1}^N {\left( {\varepsilon \log p_t^{fig} + \left( {\frac{{1 - \varepsilon }}{{N - 1}}} \right)\log p_i^{fig}} \right)}  \hfill \\
  {L_{ocr}} =  - \frac{1}{{\left| {p_{mask}^{ocr}} \right| \cdot N}}\sum\limits_{i = 1}^N {\left( {\varepsilon \log p_t^{ocr} + \left( {\frac{{1 - \varepsilon }}{{N - 1}}} \right)\log p_i^{ocr}} \right)}   
\end{flalign}

\begin{flalign}\label{eq.25}
{L_{concat}} = & - \frac{\beta }{{\left| {p_{mask}^{concat}} \right| \cdot N}}\sum\limits_{i = 1}^N {\left( {\varepsilon \log p_t^{concat} + \left( {\frac{{1 - \varepsilon }}{{N - 1}}} \right)} \right.} \nonumber \\
&\left. {\log p_i^{concat}} \right)
\end{flalign}
Here $\beta$ represents the weight value to encourage exploration. If during the training process, Mask mechanism  \uppercase\expandafter{\romannumeral3} is activated, that is, the part of the input path is masked out, it means that a new trajectory is generated, so the weight of $p^{concat}_t$ should be increased.\par

To optimize the multimodal contribution imbalance problem, the modal contribution difference ratio parameter $(\rho^{fig}_t,\rho^{ocr}_t)$ is introduced in Eq. \ref{eq.26} to adaptively adjust the gradient of each modality, where $\rho^{fig}_t$ is $\rho^{ocr}_t$. As shown in Fig.~\ref{fig:5},  the coefficient $coeff^u_n$ is integrated into the network corresponding to the modal association via Eq. \ref{eq.27} based on \cite{ref40}. At the same time, Gaussian noise $N$ is introduced to enhance the generalization ability of the model.\par
\begin{equation}\label{eq.26}
\rho _n^{ocr} = \frac{{{L_{ocr}}}}{{{L_{fig}}}}
\end{equation}
\begin{equation}\label{eq.27}
{coeff}_n^u = \begin{gathered}\begin{aligned}
\begin{cases}
  1 - \tanh \left( {\alpha  \cdot relu\left( {\rho _n^u} \right)} \right),&\for \rho _n^u > 1 \hfill \\
  1,&\for others \hfill \\
\end{cases}\end{aligned}
\end{gathered} 
\end{equation} 
\begin{flalign}
\nabla {W_{n + 1}}^u = & \nabla {W_n}^u \times coeff_n^u + N\left( {0,\sum {std\left( {\nabla {W_n}^u} \right)} } \right. \nonumber \\
& \left. { + {e^{ - 8}}} \right) 
\end{flalign}
Here $u \in \{fig,ocr\}$, and $\alpha$ is a hyperparameter that controls the degree of modulation. \par
\section{Experiments}
In this section, experiments are implemented based on two newly established datasets, and some state-of-the-art (SOTA) baselines are used for comparison. The experiment is mainly divided into four parts: link prediction main experiment, ablation experiment, training set mask experiment, and parameter interpretability analysis experiment.
\subsection{Datasets}
In this subsection, two newly established datasets, OpenBG-IMG+ and OpenBG-Complete-IMG+, are introduced.
\begin{table*}[t]
\begin{center}
\caption{Statistics of The Experimental Datasets}
\label{tab1}
\begin{tabular}{ccccccc}
\hline
Dataset    & \#Ent & \#Rel & \#Train & \#Valid & \#Test & \#num of image \\ \hline
OpenBG-IMG+ & 28891 & 136   & 197269  & 10383 & 10930  & 14718  \\
OpenBG-Complete-IMG+ & 22297 & 136 & 138479 & 7289 & 10930 & 14718 \\ \hline
\end{tabular}
\end{center}
\end{table*}
\subsubsection{OpenBG-IMG+} A new dataset OpenBG-IMG+ is created, derived from a part of the OpenBG-IMG dataset \cite{ref42}, which is a multimodal dataset in the field of e-commerce. This dataset is released in the CCKS2022 Task Three competition. Since the data set released by the competition has no correct answer, the OpenBG-IMG valid set is used as the test set in this paper, and the training set is divided into several parts as the valid set. The used dataset contains 28,891 entities and 136 relations, where only some of the head entities have image information, while none of the tail entities has image information. Each image corresponds to only one entity, and there is no duplication. Table~\ref{tab1} shows specific statistics.
\subsubsection{OpenBG-Complete-IMG+} A new repository OpenBG-Complete-IMG+ is created based on the already created database OpenBG-IMG+. Its training set and valid set are obtained from OpenBG-IMG+ deleting the triplet data with no image information in the head entity, and the test set remains unchanged. Like OpenBG-IMG+, the tail entities of all data do not contain image information, but all head entities of the training set of OpenBG-Complete-IMG+ contain image information. This new dataset contains 136 relations and 22297 entities. Table~\ref{tab1} shows specific statistics.

\subsection{Baselines}
To study the performance of IMKGA-SM, three categories of methods are used for comparison: (1) Translation-based models, TransE \cite{ref1}, TransH \cite{ref5}, and TransD \cite{ref66}. (2) Nonlinear-based models, DistMult \cite{ref22}, ComplEx \cite{ref16}, and TuckER \cite{ref7}. (3) Multimodal knowledge graph linking model, TransAE \cite{ref14}.

\subsection{Evaluation Protocol}
To further analyze the influence of image information, part of the training data is masked, and IMKGA-SM is used for experiments. According to inferences, the current dataset OpenBG-IMG+ may contain enough structural information for prediction, which interferes with the analysis of visual information. In order to highlight the role of visual information, a part of the training data is masked to create a dataset. Similar to recent works \cite{ref8} \cite{ref9}, as formalized in Eq. \ref{eq.29} and Eq. \ref{eq.30}, the mean reciprocal rank $MRR$ and the average proportion of triples with rank less than $n$ $Hits@n$ are used to evaluate inference performance.
\begin{equation}\label{eq.29}
\begin{aligned}
MRR = & \frac{1}{{\left| Q \right|}}\sum\limits_{i = 1}^{\left| Q \right|} {\frac{1}{{ran{k_i}}}}  = \frac{1}{{\left| Q \right|}}\left( {\frac{1}{{ran{k_1}}} + \frac{1}{{ran{k_2}}} +  \ldots } \right.\hfill \\
&  \left. { + \frac{1}{{ran{k_{\left| S \right|}}}}} \right) \hfill \\
\end{aligned}
\end{equation}
Here $Q$ is the set of test queries, $|Q|$ represents the number of queries, $rank_i$ is the link prediction rank of the $i$-th triple \cite{ref44}. The larger the $MRR$ indicator, the better the prediction effect.
\begin{equation}\label{eq.30}
HIT@n = \frac{1}{{\left| Q \right|}}\sum\limits_{i = 1}^{\left| Q \right|} {I\left( {ran{k_i} \leqslant n} \right)} 
\end{equation}
Here $I\left(  \cdot  \right)$ is the indicator function, if the condition is true, the function value is 1, otherwise, it is 0. The three indicators $HIT@1$, $HIT@3$, and $HIT@10$ describe the probability that the top $K(K=1,3,10)$ entities with the highest score in the link prediction contains the correct entity \cite{ref1}.
\subsection{Implementation Details} 
Next, the experiments are mainly based on the knowledge graph link prediction task. To augment the training data, each original training set triplet is reversed to generate an inverse triplet. The knowledge graph triplets in the test dataset are sorted by all entities in descending order of probability value, leaving the top ten predicted entities. Models are trained using the Adam \cite{ref45} optimizer and analyzed for hyperparameters, eigenvectors.

\subsection{Link Predict Results}

\begin{table*}[t]
\begin{center}
\caption{Results of Knowledge Graph Link Prediction on OpenBG-IMG+ and OpenBG-Complete-IMG+ Datasets}
\label{tab2}
\begin{tabular}{c|c|c|cccc|cccc}
\hline
    &   &  & \multicolumn{4}{c|}{OpenBG-IMG+}  & \multicolumn{4}{c}{OpenBG-Complete-IMG+}   \\
 Model & \!Interpretability\! & Multi-model & MRR              & Hit@1            & Hit@3            & Hit@10  & MRR              & Hit@1            & Hit@3            & Hit@10   \\ \hline
TransE \cite{ref1}    &  No  &  No   & 0.50858          & 0.33769          & 0.64876          & 0.83037   & 0.39092  &  0.24419 & 0.50155  &0.67520 \\
TransH \cite{ref5}   &  No  &  No  & 0.40132          & 0.15946          & 0.61454          & 0.81280    & 0.31544  &  0.10356 & 0.49533  &0.68197 \\
TransD \cite{ref66}   &  No  &  No  & 0.39173          & 0.15416          & 0.59158          & 0.80850  &0.30779   & 0.09954  & 0.47209  & 0.69075 \\
DistMult \cite{ref22}  &  No &  No  & 0.14469 & 0.07529    & 0.17191    & 0.35123       & 0.13496  &0.07474   & 0.16715  &   0.29560     \\
ComplEx \cite{ref16}   &  No  &  No  & 0.19756          & 0.12314          & 0.23568          & 0.38435  &  0.14017 &  0.08673 & 0.15361  & 0.31015 \\
TransAE \cite{ref14}   & \underline{No} &  \underline{Yes} & \underline{0.47107}          & \underline{0.33897}          & \underline{0.56376}          & \underline{0.75086}     & \underline{0.51174}  & \underline{0.38261}  &  \underline{0.60054} &    \underline{0.77913} \\
TuckER \cite{ref7}   & No  &  No & 0.42024          & 0.31985          & 0.49613          & 0.60795    & 0.39482  &  0.29624 & 0.48215  & 0.57557  \\\hline
IMKGA-SM(No Img) &   No &    No  &  0.64224  &  0.53284  &  0.73595  &  0.83284      &  0.59361  & 0.48948  & 0.68069  & 0.78270 \\
IMKGA-SM(MKG) &   No    &   Yes  & 0.64652      &  0.53705     &  0.73998    & 0.83678   & 0.60110  &   0.49332  & 0.68893  & 0.79780  \\
IMKGA-SM(RL) &   Yes   &   No   &  0.64341     &    0.53312   &     0.73816   &  0.83678  & 0.59582  &  0.49222  &0.68097  & 0.78948 \\
IMKGA-SM(MKG+RL) &    \textbf{Yes} &    \textbf{Yes} &  \textbf{0.64860}  & \textbf{0.53998}    &  \textbf{ 0.74089 }   &  \textbf{ 0.83714}     & \textbf{0.60177}  &  \textbf{0.49680}  &  \textbf{0.68692} &  \textbf{0.79030}\\\hline
Improv. & - & - &17.76$\%$  &20.11$\%$ & 17.72$\%$&8.63$\%$ &9.01$\%$ & 11.5$\%$ & 8.64$\%$ &1.12$\%$  \\\hline
\end{tabular}
\end{center}
\end{table*}\par
Link prediction results are shown in Table~\ref{tab2} (all scores are expressed as percentages), where the most competitive baseline TransAE \cite{ref14} results are underlined and the best results are highlighted in bold. The following points are observed.\par
Table~\ref{tab2} is studied for link prediction tasks. It is seen that the accuracy performance of IMKGA-SM is better than all other models, and IMKGA-SM uses multi-hop reasoning, which is interpretable. It is shown that the model is proven to be both interpretable and highly accurate, achieving state-of-the-art performance. The introduction of a multimodal knowledge graph in the OpenBG-IMG+ dataset is not obvious enough, so only the triplet data with image information in the head entity is retained in the new dataset. The results show that the improvement effect of IMKGA-SM in OpenBG-Complete-IMG+ is generally better than that of OpenBG-IMG+. This is speculated to be due to being overwhelmed with the help of other multimodal information when the dataset already has rich structural information.\par
To further analyze the influence of image information, a new dataset OpenBG-Complete-IMG+ is created. All head entities in the dataset have image information, and the experiment is performed again.

\subsection{Ablation Learning}
In this section, ablation learning is divided into three parts: the influence of multimodality, reward expectation, and mask mechanism on the model, and perform specific data analysis on the complete IMKGA-SM model.
\subsubsection{IMKGA-SM (No Img) vs IMKGA-SM (MKG)} To further explore the role of image information, IMKGA-SM (No image) and IMKGA-SM (MKG) are compared on two datasets, and the link prediction experiment results are shown in Table~\ref{tab2}. Image embedding and ocr embedding are added to IMKGA-SM (MKG), and a dynamic loss function adjustment mechanism is adopted. The experiment found that the improvement effect on OpenBG-Complete-IMG+ is more stable than OpenBG-IMG+.

\subsubsection{IMKGA-SM(No Img) vs IMKGA-SM(RL)} In order to verify the improvement of the model by the introduction of reward expectation, IMKGA-SM (No image) and IMKGA-SM (RL) are compared on two data sets, and the link prediction experiment results are shown in the Table~\ref{tab2}. A reward expectation mechanism based on perceptual interaction and a reward-related mask is added to IMKGA-SM (RL). The experiment results show that after adding the reward expectation mechanism $\hat R$, the model effect has been improved, but the improvement is not as much as that of MKG.

\subsubsection{IMKGA-SM (MKG+RL)} IMKGA-SM (MKG+RL) adds a masking mechanism on the basis of IMKGA-SM(MKG) and IMKGA-SM(RL) and verified its effect on OpenBG-IMG+ and OpenBG-Complete-IMG+ datasets, which are shown in Table~\ref{tab2}. Compared with the multimodal knowledge graph linking baseline TransAE \cite{ref14}, IMKGA-SM has a 17.67$\%$ improvement on the OpenBG-IMG+ dataset and a 9.01$\%$ improvement on the OpenBG-Complete-IMG+ dataset. Compared with the nonlinear-based baseline DistMult \cite{ref22}, IMKGA-SM has a 50.39$\%$ improvement on the OpenBG-IMG+ dataset and a 46.681$\%$ improvement on the OpenBG-Complete-IMG+ dataset. 

\subsection{Training Data Masking}
\begin{figure}[t]
\centering
\includegraphics[width=8cm]{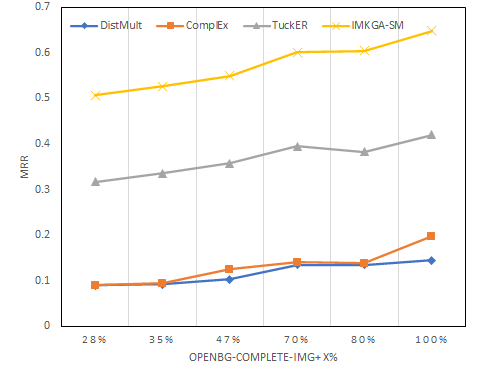}
\caption{Link prediction results (MRR) on OpenBG-Complete-IMG+x$\%$.}
\label{fig:10}
\end{figure}

\begin{figure}[t]
\centering
\includegraphics[width=8cm]{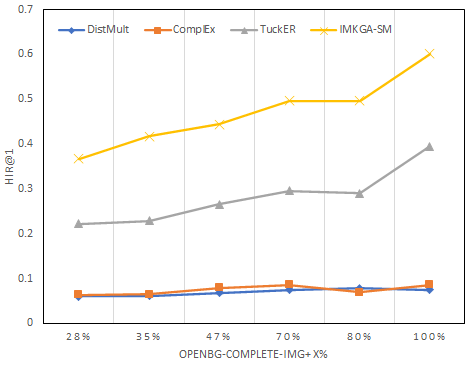}
\caption{Link prediction results (HIT@1) on OpenBG-Complete-IMG+x$\%$.}
\label{fig:11}
\end{figure}

To further explore the impact of image and structural information on the results, masking is performed on part of the training data, and IMKGA-SM is used for experiments. It is speculated that the current dataset OpenBG-IMG+ may contain enough structural information for prediction, interfering with the analysis of visual information. In order to highlight the role of visual information, a part of the training data is masked out to create datasets by controlling the frequency of a head entity with image information, which is OpenBG-IMG+28$\%$, OpenBG-IMG+35$\% $, OpenBG-IMG+47$\%$, OpenBG-IMG+70$\%$, OpenBG-IMG+80$\%$ and OpenBG-IMG+100$\%$, respectively, and the dataset information is shown in Table~\ref{tab3}. Then, the link prediction experiment is carried out again, and the results are shown in Fig.~\ref{fig:10}, 
~\ref{fig:11}. It is seen that IMKGA-SM has shown obvious advantages in data sets of different scales. The traditional method has a significant increase after adding structural information, while IMKGA-SM is still relatively stable in the improvement of the scale. IMKGA-SM not only has comparable generalization ability to neural network models, but also has stronger interpretability than other baseline methods.\par

\begin{table}[t]
\begin{center}
\caption{Statistics of Datasets in Training Data Masking Experience}
\label{tab3}
\begin{tabular}{ccccccc}
\hline
Dataset    & \#Ent & \#Rel & \#Train & \#Valid & \#Test \\ \hline
OpenBG-IMG+80\% &27839 &136 &158527 &8344 &10930 \\
OpenBG-IMG+70\% & 22297 & 136 & 138479 & 7289 & 10930  \\ 
OpenBG-IMG+47\% &21817 & 136&92648 &4877 & 10930 \\
OpenBG-IMG+35\% &21469 &136 &68599 &3611 &10930 \\
OpenBG-IMG+28\% &21215 &136 &55397 & 2916& 10930 \\\hline
\end{tabular}
\end{center}
\end{table}\par

\par
\par

\subsection{Parameter Interpretability}
In this subsection, the parameter interpretability is divided into three parts, mainly analyzing the impact of different batch sizes, label smooth and modulation impact.
\subsubsection{The Influence of Different Batch Sizes $N$} 
\begin{figure}[t]
\centering
\includegraphics[width=8cm]{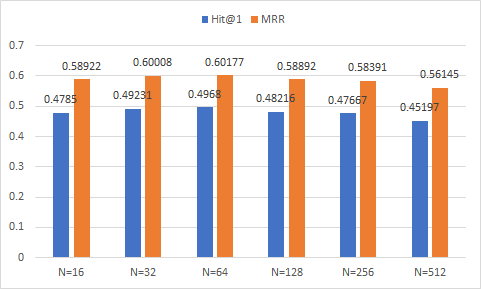}
\caption{Link prediction results on OpenBG-Complete-IMG+ in different $N$.}
\label{fig:20}
\end{figure}

Fig.~\ref{fig:20} investigates the effect of different batch sizes $N$. The batch size is set to $N \in[16,32,64,128,256,512]$. It is observed that as $N$ increases, the performance of IMKGA-SM rises first and then declines steadily in most cases, presumably because undertraining and overfitting negatively affect the model. The results show that an appropriate training parameter size improves the effectiveness of the inference model. From the experimental results, the optimal parameter is $N = 16$.\par
\par
\subsubsection{The Influence of Different Labels Smooth $\varepsilon$} 
\begin{figure}[t]
\centering
\includegraphics[width=8cm]{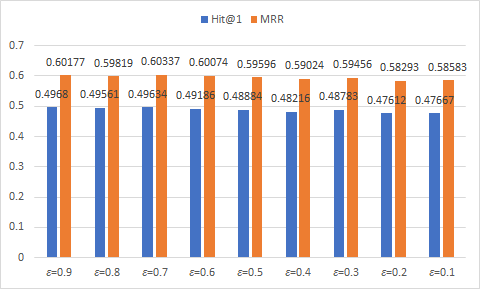}
\caption{Link prediction results on OpenBG-Complete-IMG+ in different $\varepsilon$.}
\label{fig:e}
\end{figure}

In this paper, the method of label smoothing is used in the loss function, which is a regularization method to prevent overfitting. By setting $\alpha=0.5$, the effect of the label smoothing parameter $\varepsilon$ is shown in Fig.~\ref{fig:e}. The batch size is set to $\varepsilon \in[0.1,0.2,0.3,0.4,0.5,0.6,0.7,0.8,0.9]$. It is observed that with the decrease of $\varepsilon$, the performance of IMKGA-SM rises first and then declines steadily in most cases, which is probably because MRR is the main evaluation indicator, and the shrinking of its proportion has a negative effect on the final result influences. The results show that the optimal parameter is $\varepsilon = 0.7$.\par
\par
\subsubsection{The Influence of Different Modulation Impact $\alpha$} 
\begin{figure}[t]
\centering
\includegraphics[width=8cm]{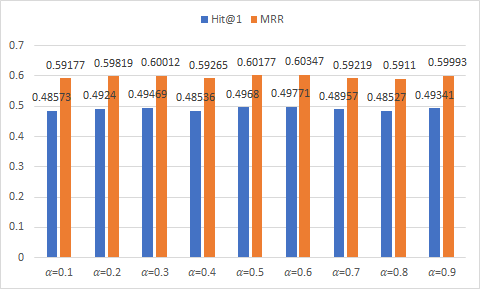}
\caption{Link prediction results on OpenBG-Complete-IMG+ in different $\alpha$.}
\label{fig:a}
\end{figure}

With $\varepsilon=0.9$, the effect of the modulation impact $\alpha$ on IMKGA-SM is demonstrated in Fig.~\ref{fig:a}. From the results, $\alpha =0.6$ is observed to be the optimal value on the data set.\par
\par

\section{Conclusions and Discussion}
In this paper, how to effectively utilize multi-modal auxiliary features for multi-hop knowledge graph inference is investigated, aiming to improve the accuracy of model inference and achieve interpretability simultaneously. An efficient IMKGA-SM model is proposed, which outperforms existing methods on the multimodal knowledge graph inference task. In IMKGA-SM, structural features and multimodal data are first extracted in-depth, and then a return-to-go mechanism based on perceptual similarity is constructed and applied to the large sequence framework. In addition, three mask mechanisms are designed to alleviate the problem of data sparsity. Next, a multimodal autoregressive loss function adjustment mechanism is introduced to take full advantage of multimodality. Finally, experimental results show that IMKGA-SM achieves higher effectiveness and interpretable ability versus other trending rivals in knowledge graph link prediction tasks. To conclude, IMKGA-SM requires effective methods to minimize the negative impact of sparse data. These tasks are left for future work.

%\section*{Acknowledgment}
%The authors would like to thank the anonymous reviewers for their constructive comments. This work is supported in ...

\bibliographystyle{IEEEtran}
\bibliography{tkderef}


%\end{IEEEbiography}


%\newpage



%\vfill


\end{document}